\documentclass[conference]{IEEEtran}
\IEEEoverridecommandlockouts
\usepackage{cite}
\usepackage{amsmath,amssymb,amsfonts}
\usepackage{algorithmic}
\usepackage{graphicx}
\usepackage{textcomp}
\usepackage{cuted}
\usepackage{booktabs}
\usepackage{multirow}
\newcommand{\equalcontrib}{\textsuperscript{*}}

\def\BibTeX{{\rm B\kern-.05em{\sc i\kern-.025em b}\kern-.08em
    T\kern-.1667em\lower.7ex\hbox{E}\kern-.125emX}}
\begin{document}

\title{Residual-based Efficient Bidirectional Diffusion Model for Image Dehazing and Haze Generation}

\author {
    % Authors
    Bing Liu\textsuperscript{\rm 1,\rm 2}\equalcontrib\thanks{* These authors contributed equally to this work.},
    Le Wang\textsuperscript{\rm 1,\rm 2}\equalcontrib\thanks{Corresponding author. Email: TS23170132P31@cumt.edu.cn},
    Hao Liu\textsuperscript{\rm 1,\rm 2},
    Mingming Liu\textsuperscript{\rm 1}\\
    \textsuperscript{\rm 1}China University of Mining and Technology, Xuzhou, Jiangsu, China\\
    \textsuperscript{\rm 2}Mine Digitization Engineering Research Center of the Ministry of Education, Xuzhou, Jiangsu, China\\
    liubing@cumt.edu.cn, TS23170132P31@cumt.edu.cn, tb20170007b4@cumt.edu.cn, mingming.liu@cumt.edu.cn
}

\maketitle

\begin{abstract}
Current deep dehazing methods only focus on removing haze from hazy images, lacking the capability to translate between hazy and haze-free images. To address this issue, we propose a residual-based efficient bidirectional diffusion model (RBDM) that can model the conditional distributions for both dehazing and haze generation. Firstly, we devise dual Markov chains that can effectively shift the residuals and facilitate bidirectional smooth transitions between them. Secondly, the RBDM perturbs the hazy and haze-free images at individual timesteps and predicts the noise in the perturbed data to simultaneously learn the conditional distributions. Finally, to enhance performance on relatively small datasets and reduce computational costs, our method introduces a unified score function learned on image patches instead of entire images. Our RBDM successfully implements size-agnostic bidirectional transitions between haze-free and hazy images with only 15 sampling steps. Extensive experiments demonstrate that the proposed method achieves superior or at least comparable performance to state-of-the-art methods on both synthetic and real-world datasets. 
\end{abstract}


\section{Introduction}
\label{sec:intro}

Due to the absorption and scattering of light during propagation in adverse weather conditions, photographs captured in hazy weather typically exhibit low contrast and fidelity \cite{c:1}. The process of enhancing the contrast and color integrity of images captured in hazy environments is known as haze removal or dehazing \cite{c:2}. The hazing process is formulated by the well-known atmospheric scattering model (ASM) \cite{c:3}:
\begin{equation}
I(p) = J(p) e^{-\beta d(p)} + A (1 - e^{-\beta d(p)}), \label{eq:1}
\end{equation}
where \(p\) is the pixel position, \(I(p)\) and \(J(p)\) are the hazy and haze-free images respectively. A denotes the global atmospheric light. \(e^{-\beta d(p)}\) is the transmission map, in which \(d(p)\) is the scene depth and the scattering coefficient \(\beta\) represents the degree of haze. However, the presence of various possible haze densities and lighting conditions under hazy conditions leads to the  ill-posedness of this problem.

 % With the advent of deep learning, significant advancements have been made in image dehazing.

Early image dehazing methods \cite{c:4,c:5,c:6,c:7} rely on various image priors obtained from the observation and statistical analysis of natural hazy images to estimate \(A\) and \(e^{-\beta d(p)}\). Nevertheless, these methods exhibit limited robustness. Current deep neural network-based methods can be broadly categorized into two types: those dependent on physical models and those that directly learn the haze-free counterpart. The former approach estimates the transmission map and global atmospheric light, then uses an inverse physical scattering model to restore the clear image \cite{c:8,c:9,c:10}. The latter approach investigates an end-to-end mapping from hazy images to their haze-free equivalents \cite{c:11,c:12,c:13,c:14,c:15}. Despite these advancements, the inherent limitations of DNNs, which are typically unidirectional, hinder their adaptability and necessitate model retraining for each new scenario.

\begin{figure}[t]
\centering
% \vspace{-50pt}
\includegraphics[width=\linewidth]{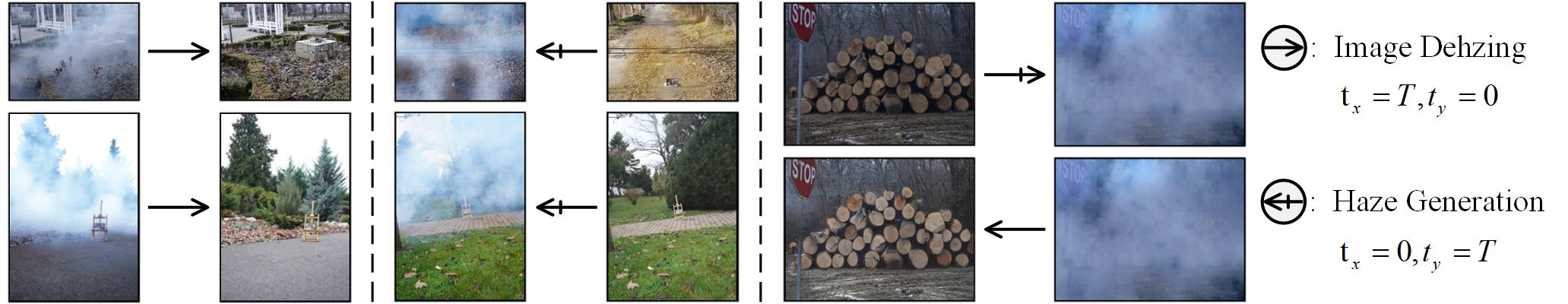}
\caption{Given any two unordered image collections clear \(X\) and hazy \(Y\), our RBDM can perform both dehazing and haze generation. \textit{(left)} Dehazing hazy images. \textit{(center)} Hazing haze-free images. \textit{(right)} Hazing haze-free images followed by dehazing.}
\label{fig:1}
\end{figure}

Recently, diffusion models \cite{c:16,c:17,c:18,c:19,c:20,c:21} have demonstrated strong ability in modeling data distributions, show potential for addressing downstream low-level vision tasks. Although current diffusion-based dehazing models use hazy images as priors and leverage skip connections to utilize information from hazy images \cite{c:21}, how to perform a more efficient and interpretable forward process by starting from a prior distribution based on the hazy image needs to be further explored. 

To incorporate the prior about the hazy image and improve the generalization performance, we propose a Residual-based Efficient Bidirectional Diffusion Model for Image Dehazing and Haze Generation (RBDM). This model performs both image dehazing and haze generation by setting different timesteps \(t_{x}\) and \(t_{y}\) as shown in Fig.~\ref{fig:1}. Our approach is based on the convergence of the initial state of the Markov chain to an approximate distribution of the haze-free image and the final state to an approximate distribution of the hazy image for dehazing tasks. In haze generation, the initial and final states of the Markov chain are inverted in contrast to the dehazing process.This dual convergence enables the reconstruction of haze-free images from hazy ones and vice versa, leading to a single diffusion model with shorter Markov chains. Additionally, we incorporate a patch-based diffusion model \cite{c:20} to enable size-agnostic processing and reduce computational complexity. Our RBDM outperforms existing diffusion-based dehazing methods due to its efficient propagation of residual information over multiple steps and its ability to effectively perform bidirectional generation between hazy and haze-free images. Furthermore, our design allows for a concise expression of the evidence lower bound, simplifying the training process. The contributions of this paper are summarized as follows:
\begin{itemize}
\item We propose a residual-based diffusion model for image dehazing, which constructs a Markov chain that transfers between the hazy image and the haze-free image by shifting their residuals, aiming to precisely reconstruct the haze-free image with a shorter chain.
\item The proposed RBDM models the conditional distributions for both dehazing and haze generation to realize efficient bidirectional transitions, which can not only maintain high-quality dehazing, but also facilitate data augmentation with realistic haze synthesis.
\item We implement the proposed RBDM and extensively evaluate the proposed RBDM on widely-used benchmark datasets. The qualitative and quantitative experiments demonstrate its superior performance compared to state-of-the-art methods. 
\end{itemize}

\section{RELATED WORK}
\noindent\textbf{Single Image Dehazing.} Existing image dehazing methods can be classified into two categories: prior-based methods, which utilize empirical priors derived from the statistical differences between hazy and haze-free images, and learning-based methods, which learn mapping functions in a data-driven manner.

Prior-based methods, such as the atmospheric scattering model (ASM) \cite{c:3} and handcrafted priors like dark channel prior (DCP) \cite{c:6}, non-local prior (NLP) \cite{c:22}, and color attenuation prior (CAP) \cite{c:7}, rely on statistical differences between hazy and haze-free images. While effective, their performance is limited by the assumptions underlying these priors.

Learning-based methods, utilizing deep learning, aim to learn data-driven mappings for dehazing \cite{c:24,c:11,c:9,c:13,c:10}. Early CNN-based approaches estimate transmission maps and global atmospheric light \cite{c:11}, while more recent models like AOD-Net \cite{c:9} generate haze-free images directly. Although unsupervised methods such as Cycle-Dehaze \cite{c:26} improve generalization to real-world scenarios, they still struggle with non-homogeneous haze. Methods like UDN \cite{c:18} address uncertainties but do not fully resolve non-homogeneous haze.

\section{METHODOLOGY}
\begin{figure*}[t]
\centering
\includegraphics[width=1\textwidth]{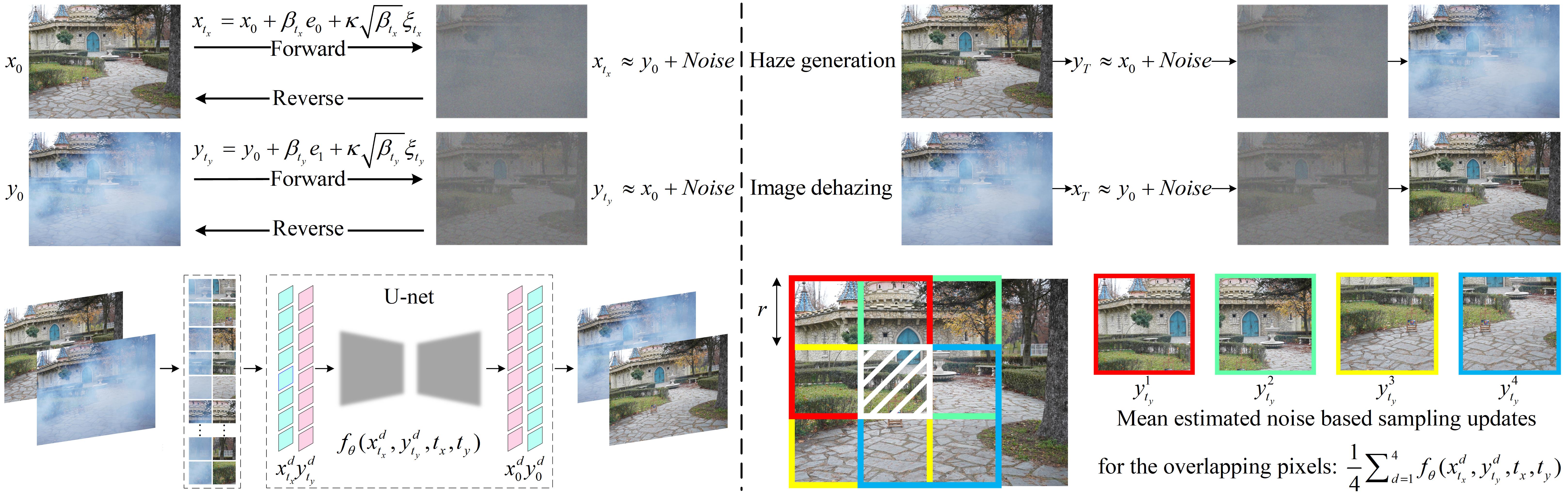}
\caption{\textbf{Overview of the proposed RBDM. } \textit{(left)} Illustration of residual-based efficient bidirectional diffusion model. \textit{(right)} Illustration of overlapping pixel updates during patch-based sampling. We show a simplified scenario where the overlap factor \(r = p/2\), involving four intersecting patches that share a grid cell highlighted with white stripes. In this setup, updates to pixel sampling are performed by averaging the estimated noise from the four intersecting patches at each step of the denoising process. }
\label{fig:2}
\end{figure*}

Hazy and haze-free images belong to different domains but share common features.
To optimally maintain these shared features, the proposed RBDM constructs two Markov chains that serve as the bridge between the haze and haze-free images as shown in Fig.~\ref{fig:2}. For conciseness, the haze and haze-free images are denoted as \(y_{0}\) and \(x_{0}\) respectively. 
\subsection{Model Design}
% Let's briefly review diffusion models\cite{c:16,c:17,c:18}. The core idea of diffusion models, which are essentially probabilistic models for generating data, is to generate new data points by simulating the perturbation process of data. Diffusion models consist of two main processes: the forward process and the reverse process. In the forward process, the model gradually adds noise to the data, enabling the data distribution to gradually approximate a simple prior distribution (usually a Gaussian distribution). In the reverse process, the model starts from a simple prior distribution (such as a Gaussian distribution) and gradually removes the noise, attempting to recover the original data.

Different from existing DDPM methods, which generates data starting from pure Gaussian noise.
We sample from the approximate distribution of hazy images to generate haze-free images, $\textit i.e.$, image dehazing, and from the approximate distribution of haze-free images to generate hazy images, $\textit i.e.$, haze generation.

\noindent\textbf{Forward Process.} We define the residual between hazy and haze-free images as \(e_{0}\), i.e., \(e_{0}=y_{0}-x_{0}\).
The core idea is to start from a haze-free image \(x_{0}\) and gradually adjust the residual \(e_{0}\) while injecting noise through a Markov chain with length \(T\), with the final state converging to an approximate distribution of the hazy image \(y_{0}\). The corresponding Markov chain is formulated as:
\begin{equation}
\begin{split}
q(x_{t_{x}}|x_{t_{x}-1},y_{t_{y}} )=N(x_{t_{x}};x_{t_{x}-1}+\alpha _{t_{x}}e_{0},\kappa ^{2}\alpha _{t_{x}} I), 
\end{split}
\label{eq:2}
\end{equation}
where \(t_{x}\) and \(t_{y}\) are two timesteps that can be different, and 
\(x_{t_{x}}\) and \(y_{t_{y}}\) are the corresponding perturbed data. Note that \(\alpha_{t_{x}}=\beta_{t_{x}}-\beta_{t_{x}-1}\) for \(t_{x}>1\) and \(\alpha_{1}=\beta_{1}\), a sequence of hyperparameters \(\left\{{ \beta_{t} }\right \}_{t=1}^{T}\) controls the transition speed from the initial to the final state, increasing monotonically such that \(\beta_{1}\longrightarrow0\) and \(\beta_{T}\longrightarrow 1 \).
The parameter \(\kappa\) regulates the noise variance, ensuring a smooth transition between \(x_{t}\) and \(x_{t-1}\) through \(\kappa ^{2}\alpha _{t_{x}}\), considering the principle of standard deviation\cite{c:34}, where \(I\) is the identity matrix. Notably, in image dehazing, setting \(t_{x}=T\) and \(t_{y}=0\) is typically applied. Crucially, the marginal distribution at any timestep \(t_{x}\) is analytically integrable:
\begin{equation}
q(x_{t_{x}}|x_{0},y_{t_{y}} )=N(x_{t_{x}};x_{0}+\beta _{t_{x}}e_{0},\kappa ^{2}\beta _{t_{x}} I),
\label{eq:3}
\end{equation}
As illustrated, the marginal distributions of \(x_{1}\) and \(x_{T}\) converge to \(\delta _{x_{0}}(\cdot )\) and \(N(\cdot ;y_{0},\kappa ^{2}I)\) respectively. In other words, by setting the transition distribution according to Eq.\eqref{eq:2}, the initial state in the dehazing task approximates the distribution of haze-free images, while the final state approximates the distribution of hazy images. This allows the model to effectively learn to reconstruct haze-free images from hazy ones.

Since haze generation differs from dehazing, we define the residual  \(e_{1}=x_{0}-y_{0}\) and perturb \(y_{0}\) by shifting \(e_{1}\) through a Markov chain of length \(T\), as shown below:
\begin{equation}
q(y_{t_{y}}|y_{t_{y}-1},x_{t_{x}} )=N(y_{t_{y}};y_{t_{y}-1}+\alpha _{t_{y}}e_{1},\kappa ^{2}\alpha _{t_{y}} I),
\label{eq:4}
\end{equation}
Likewise the marginal distribution at any time \(t_{y}\) is defined as:
\begin{equation}
q(y_{t_{y}}|y_{0},x_{t_{x}} )=N(y_{t_{y}};y_{0}+\beta _{t_{y}}e_{1},\kappa ^{2}\beta _{t_{y}} I).
\label{eq:5}
\end{equation}
In this process, all parameters remain the same as in the setting of image dehazing, except that it specifically applies the settings \(t_{x}=0\) and \(t_{y}=T\). The marginal distributions of \(y_{1}\) and \(y_{T}\) converge to \(\delta _{y_{0}}(\cdot )\) and \(N(\cdot ;x_{0},\kappa ^{2}I)\) respectively. 

\noindent\textbf{Reverse Process.} The reverse processes aim to estimate the posterior distributions \(p(x_{0}|y_{t_{y}})\) and \(p(y_{0}|x_{t_{x}})\) for all \(t_{x}\) and \(t_{y}\), respectively. The generation processes are defined as follows:
\begin{equation}
\begin{aligned}
p(x_{0}|y_{t_{y}}) \! = \! \! \int \!p(x_{T}|y_{t_{y}})\! \prod_{t_{x}=1}^{T} \!p_{\theta}(x_{t_{x}-1}|x_{t_{x}},y_{t_{y}})dx_{1:T}, \\
p(y_{0}|x_{t_{x}})\! = \! \! \int \! p(y_{T}|x_{t_{x}}) \! \prod_{t_{y}=1}^{T} \!p_{\theta}(y_{t_{y}-1}|y_{t_{y}},x_{t_{x}})dy_{1:T},
\end{aligned}
\label{eq:6_and_7}
\end{equation}
where $p(x_{T}|y_{t_{y}})\! \approx \!N(x_{T}|y_{t_{y}},\kappa ^{2}I)$, $p_{\theta}(x_{t_{x}-1}|x_{t_{x}},y_{t_{y}})$ is the inverse transition kernel from \(x_{t_{x}}\)to \(x_{t_{x}-1}\) with a learnable parameter \(\theta\). Similarly, $p(y_{T}|x_{t_{x}})\! \approx \!N(y_{T}|x_{t_{x}},\kappa ^{2}I)$,
\(p_{\theta}(y_{t_{y}-1}|y_{t_{y}},x_{t_{x}})\) is the inverse transition kernel from \(y_{t_{y}}\) to \(y_{t_{y}-1}\). 
According to Bayes’s theorem, we adopt the assumptions : 
\begin{equation}
\begin{aligned}
p_{\theta}(x_{t_{x}-1}|x_{t_{x}},y_{t_{y}})&=N(x_{t_{x}-1};\mu_{\theta }^{x}(x_{t_{x}},y_{t_{y}},t_{x},t_{y}),\Sigma^{x}_{\theta}I), \\
p_{\theta}(y_{t_{y}-1}|y_{t_{y}},x_{t_{x}})&=N(y_{t_{y}-1};\mu_{\theta }^{y}(x_{t_{x}},y_{t_{y}},t_{x},t_{y}),\Sigma^{y}_{\theta}I),
\end{aligned}
\end{equation}
The optimization for \(\theta\) is performed by minimizing the following negative evidence lower bounds,
\begin{equation}
\hspace{-0.48em}
\begin{aligned}
\!\min_{\theta }\!\!\sum_{t_{x}}\!D_{KL}[q(x_{t_{x}-1}|x_{t_{x}},x_{0},y_{t_{y}})||p_{\theta}(x_{t_{x}-1}|x_{t_{x},}y_{t_{y}})],&\\
\!\min_{\theta }\!\!\sum_{t_{y}}\!D_{KL}[q(y_{t_{y}-1}|y_{t_{y}},y_{0},x_{t_{x}})||p_{\theta}(y_{t_{y}-1}|y_{t_{y},}x_{t_{x}})],& 
\end{aligned}
\label{eq:8_and_9}
\end{equation}
where \(D_{KL}[\cdot||\cdot]\) denotes the Kullback-Leibler (KL) divergence. More mathematical details can refer to \cite{c:18} or \cite{c:16}.

\begin{table*}[t]
    \caption{Quantitative comparison on the synthetic dataset. The best results are in \textbf{bold}, and the second best are with \underline{underline}.}
    \centering
    % \resizebox{.95\columnwidth}{!}{
    \begin{tabular}{llccccccc}
        \toprule
        Datasets & Indicators & DCP \cite{c:6}  & AOD-Net \cite{c:39}  & FFA-Net \cite{c:14} & DehazeFormer \cite{c:42} & MixDehazeNet \cite{c:43} & TSNet \cite{c:44} & Ours \\
        \midrule
        \multirow{2}{*}{RESIDE-6K} 
        & PSNR $\uparrow$    & 17.88  & 20.27      & 29.96    & 30.89             & 30.69 & \underline{31.31} &\textbf{32.68} \\
        & SSIM $\uparrow$    & 0.816  & 0.855      & 0.973    & \textbf{0.977} & 0.974 & 0.975             &\underline{0.976} \\
        \bottomrule
    \end{tabular}
    \label{table:1}
\end{table*}

\begin{table*}[t]
\caption{Quantitative comparisons for non-homogeneous dehazing on NHIRE2020, NHIRE2021, and NHIRE2023 datasets. The best
results are in \textbf{bold},and the second best are with \underline{underline}. }
\centering
\begin{tabular}{lcccccccc}
    \toprule
    \textbf{Methods} & \multicolumn{2}{c}{NTIRE2020} & \multicolumn{2}{c}{NTIRE2021} & \multicolumn{2}{c}{NTIRE2023} & \multicolumn{2}{c}{Average} \\
    \cmidrule(lr){2-3} \cmidrule(lr){4-5} \cmidrule(lr){6-7} \cmidrule(lr){8-9}
    & PSNR$\uparrow$ & SSIM$\uparrow$ & PSNR$\uparrow$ & SSIM$\uparrow$ & PSNR$\uparrow$ & SSIM$\uparrow$ & PSNR$\uparrow$ & SSIM$\uparrow$ \\
    \midrule
    Hazy           & 11.31 & 0.4160 & 11.24 & 0.5787 & 8.86 & 0.4702 & 10.47 & 0.4883 \\
    DCP (TPAMI’10)  & 12.35 & 0.4480 & 10.57 & 0.6030 & 10.98 & 0.4777 & 11.30 & 0.5096 \\
   AODNet (ICCV’17)   & 14.04 & 0.4450 & 14.52 & 0.6740 & 13.75 & 0.5619 & 14.10 & 0.5603
 \\
   GridDehazeNet (ICCV’19) & 14.78 & 0.5074 & 18.05 & 0.7433 & 16.85 & 0.6075 & 16.56 & 0.6194 \\
    FFANet (AAAI’20) & 16.98 & 0.6105 & 19.75 & \underline{0.7925} & 17.85 & 0.6485 & 18.20 & 0.6838  \\
    TNN (CVPRW’21) & 17.18 & 0.6114 & 20.13 & \textbf{0.8019} & 18.19 & 0.6426 & 18.50 & 0.6853\\
    DeHamer (CVPR’22) & 18.53 & 0.6201 & 18.17 & 0.7677 & 17.61 & 0.6051 & 18.10 & 0.6693 \\
    SCANet (CVPR’23)  & \underline{19.52} & \underline{0.6488} & \underline{21.14} & 0.7694 & \underline{20.44} & \underline{0.6616} &  \underline{20.37} & \underline{0.6933}\\
    \midrule
    Ours & \textbf{23.19} & \textbf{0.6665} & \textbf{24.84} & 0.7563 & \textbf{23.65} & \textbf{0.6760} & \textbf{23.89} & \textbf{0.7085}  \\
    \bottomrule
\end{tabular}
\label{table:2}
\end{table*}

Based on Eq.\eqref{eq:2} and Eq.\eqref{eq:3}, the target distribution \(q(x_{t_{x}-1}|x_{t_{x}},x_{0},y_{t_{y}})\) and \(q(y_{t_{y}-1}|y_{t_{y}},y_{0},x_{t_{x}})\) in Eq.\eqref{eq:6_and_7} can be tractable and expressed in an explicit form as follows:
\begin{equation}
\begin{aligned}
q(x_{t_{x}-1}|x_{t_{x}},x_{0},y_{t_{y}}) = N(x_{t_{x}-1}|\frac{\beta _{t_{x}-1}}{\beta_{t_{x}}}+\frac{\alpha _{t_{x}}}{\beta_{t_{x}}}x_{0},\\
\kappa ^{2}\frac{\beta _{t_{x}-1}}{\beta_{t_{x}}}\alpha_{t_{x}}I) ,\\
q(y_{t_{y}-1}|y_{t_{y}},y_{0},x_{t_{x}}) = N( y_{t_{y}-1}|\frac{\beta _{t_{y}-1}}{\beta_{t_{y}}}+\frac{\alpha _{t_{y}}}{\beta_{t_{y}}}y_{0},\\
\kappa ^{2}\frac{\beta _{t_{y}-1}}{\beta_{t_{y}}}\alpha_{t_{y}}I) ,
\label{eq:10_and_11}
\end{aligned}
\end{equation}

\noindent Considering that the variance parameters are independent of \(x_{t_{x}}\) and
\(y_{t_{y}}\), we set:
\begin{equation}
\begin{aligned}
\Sigma_{\theta}^{x}(x_{t_{x}},y_{t_{y}},t_{x},t_{y})=\kappa ^{2}\frac{\beta _{t_{x}-1}}{\beta_{t_{x}}}\alpha_{t_{x}}I,\\
\Sigma_{\theta}^{y}(x_{t_{x}},y_{t_{y}},t_{x},t_{y})=\kappa ^{2}\frac{\beta _{t_{y}-1}}{\beta_{t_{y}}}\alpha_{t_{y}}I,
\end{aligned}
\end{equation}
In addition, the predicted mean values \(\mu_{\theta }^{x}(x_{t_{x}},y_{t_{y}},t_{x},t_{y})\) and \(\mu_{\theta }^{y}(x_{t_{x}},y_{t_{y}},t_{x},t_{y})\) of the noise are reparameterized as follows:
\begin{equation}
\begin{aligned}
\mu_{\theta }^{x}(x_{t_{x}},y_{t_{y}},t_{x},t_{y})= \frac{\beta _{t_{x}-1}}{\beta_{t_{x}}}x_{t_{x}} +\frac{\alpha _{t_{x}}}{\beta _{t_{x}}}f_{\theta }(x_{t_{x}},y_{t_{y}},t_{x},t_{y}),\\
\mu_{\theta }^{y}(x_{t_{x}},y_{t_{y}},t_{x},t_{y})= \frac{\beta _{t_{y}-1}}{\beta_{t_{y}}}y_{t_{y}} +\frac{\alpha _{t_{y}}}{\beta _{t_{y}}}f_{\theta }(x_{t_{x}},y_{t_{y}},t_{x},t_{y}),
\label{eq:11}
\end{aligned}
\end{equation}
where \(f_{\theta}\) is a deep neural network with the parameter \(\theta\). Inspired by the unified view, we no longer differentiate between networks for dehazing and haze generation processes. 

Based on Eq.\eqref{eq:11}, we simplify the objective functions as follows,
\begin{equation}
\begin{aligned}
\min_{\theta}\sum \omega ||f_{\theta}(x_{t_{x}},y_{t_{y}},t_{x},t_{y})-(x_{0},y_{0})||_{1},
\end{aligned}
\end{equation}
where \(\omega=1\), consistent with the conclusions of \cite{c:16}, we find that discarding the weight \(\omega\) results in a more evident improvement in performance.

\subsection{Patch Diffusion Training and Inference}
Diffusion models typically perform global modeling with a fixed input resolution and are moved into latent space to reduce computational complexity. Thus, not only does performance degrade with mismatched input sizes, but mapping high-dimensional features to a lower-dimensional space also results in the loss of fine details and textures. To address the above issues, we employ patch-based diffusion \cite{c:35,c:36}.

During training, we randomly generate a binary mask matrix \(P_{i}\) of the same dimensionality as \(x_{0}\) and \(y_{0}\), indicating the location of the i-th \(p \times p\) patch from the image. Then, by setting \(x^{i}_{0} = Crop(P_{i} \circ X_{0}) \) and \(y^{i}_{0} = Crop(P_{i} \circ Y_{0})\) , we obtain \(p \times p\) patchs from a training image pair \((x_{0},y_{0})\), where the \(Crop(\cdot)\) operation extracts the patch from the location indicated by \(P_{i}\).  

During inference, we first decompose an input image of arbitrary size into a dictionary \(B\) of overlapping patches by moving a window of size \(p \times p\) with a stride of \(r\) in both horizontal and vertical dimensions as shown in Fig.~\ref{fig:2} \textit{(right)}. Then, at each sampling time step \(t\): use noise estimator network \(f_{\theta }(x^{b}_{t_{x}},y^{b}_{t_{y}},t_{x},t_{y})\)
to directly estimate the clean image for each patch, accumulate these overlapping results in a matrix \(Sample\) of the same size as the entire image, and use a mask matrix $M$ of the same size to record the overlapping regions. At the end of the current sampling time step, take the average of the overlapping pixel parts based on the mask to obtain the final output image.

By processing smaller, fixed-size regions independently, we mitigate input size mismatch and preserve fine details, avoiding latent space compression.

\begin{figure}[htbp]
\centering
% \vspace{-50pt}
\includegraphics[width=\linewidth]{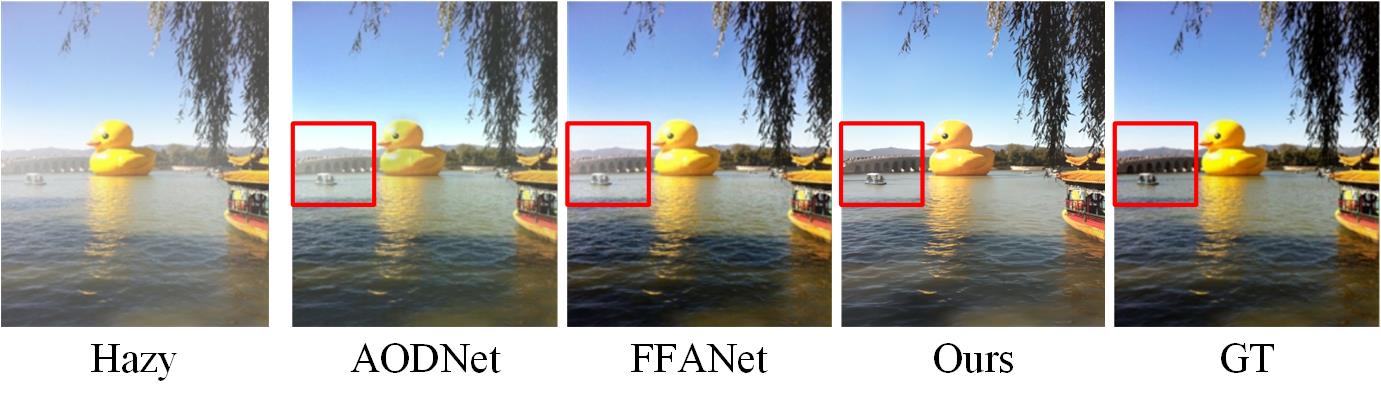}
\caption{Visual comparisons of various methods on RESIDE-6K dataset.}
\label{fig:3}
\end{figure}

\begin{figure*}[!h]
\centering
\includegraphics[width=1\textwidth]{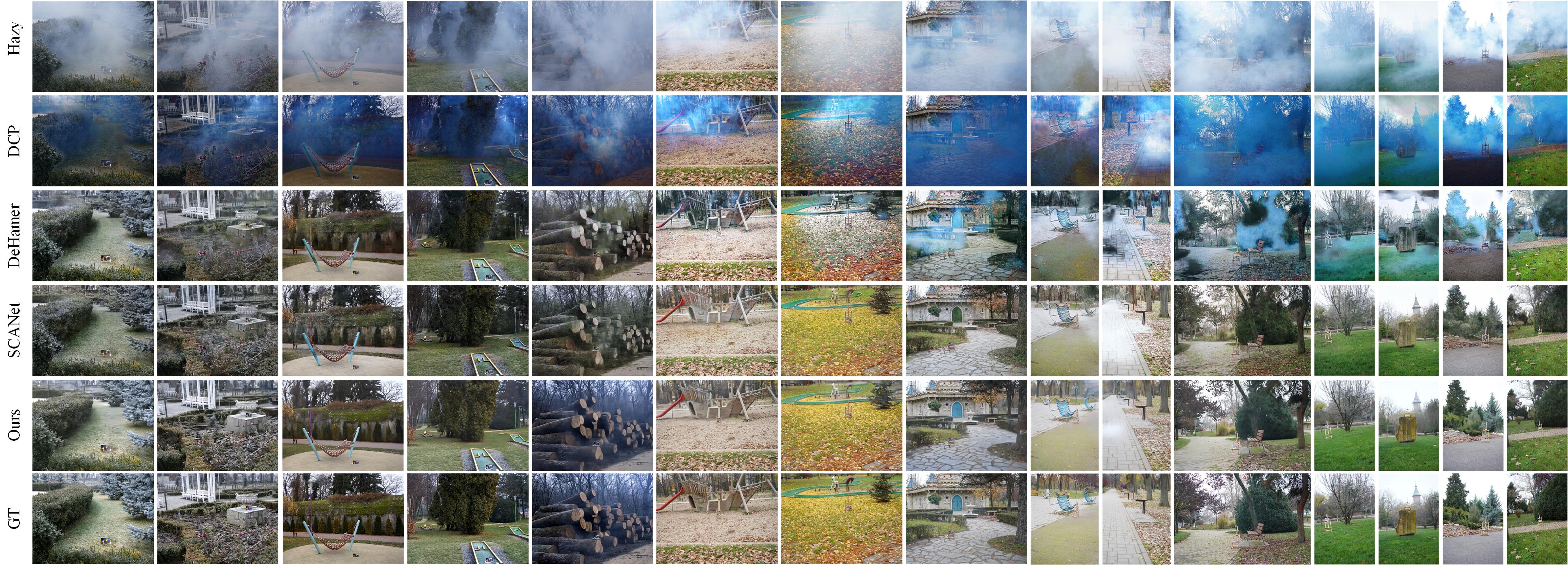} % Reduce the figure size so that it is slightly narrower than the column.
\caption{Visual comparisons of various methods on NTIRE2020, NTIRE2021, and NTIRE2023 datasets.
}
\label{fig:4}
\end{figure*}

\section{EXPERIMENTS}
\subsection{Datasets}

We train and evaluate the proposed method on publicly available datasets RESIDE-6K\cite{c:37}, NTIRE2020\cite{c:45,c:46}, NTIRE2021\cite{c:47} and NTIRE2023\cite{c:48}. RESIDE-6K contains 3,000 pairs of synthetic indoor images and 3,000 pairs of outdoor images, with 5,000 pairs used for training and 1,000 pairs for testing. NTIRE2020, NTIRE2021, NTIRE2023 datasets exhibit haze patterns that are non-uniformly distributed. Specifically, NTIRE2020
contains 45 training, 5 validation, and 5 test image pairs. NTIRE2021 contains 25 training, 5 validation, and 5 test image pairs. NTIRE2023 contains 40 training, 5 validation, and 5 test image pairs.

\subsection{Training Details}
The proposed method is implemented by PyTorch 1.13.1 and trained on a PC Intel(R) Core(TM) i5-13600K CPU @5.10GHz and NVIDIA GeForce RTX 4090 GPU. The optimizer used is Adam with a fixed learning rate of 5e-5. The network architecture is based on a modified version of the UNet from DDPM \cite{c:16}, tailored to suit our specific task. Our RBDM is trained on the RESIDE-6K dataset for 190K iterations. Specifically, we initially sample 16 images from the training set and randomly extracted 16 patches of size 64 \(\times\) 64 from each image, resulting in mini-batches of 256 patches. For the NTIRE2020, NTIRE2021, and NTIRE2023 datasets, 8 image pairs are randomly chosen, with each image being cropped into 8 patches of size 128 \(\times\) 128, resulting in mini-batches of 64 patches and the model is trained on this dataset for a total of 900K iterations.

\subsection{Comparison with state-of-the-art methods} 

As shown in Table~\ref{table:1}, our RBDM achieves the highest PSNR and SSIM scores on RESIDE-6K, demonstrating superior performance in both haze removal and the preservation of image details and textures. As shown in Fig.~\ref{fig:3}, traditional methods often lead to over-enhanced colors, resulting in unnatural outputs. In contrast, our RBDM maintains the natural color of the images while effectively removing haze, avoiding the artificial look caused by excessive processing. 
In summary, RBDM strikes an optimal balance between haze removal, detail preservation, and natural color rendering, producing more realistic and visually appealing dehazed images.

\begin{table}[t]
  \caption{Performance comparison of RBDM on the NTIRE2020 under different configurations.}
  \centering
  \begin{tabular}{ccccc}
    \toprule
    \multicolumn{3}{c}{Configurations} & \multicolumn{2}{c}{Metrics} \\
    \cmidrule(lr){1-3} \cmidrule(lr){4-5}
    \textbf{timesteps} & \textbf{variance factor $\kappa$} & \textbf{overlap factor $r$} & \textbf{PSNR$\uparrow$} & \textbf{SSIM$\uparrow$} \\
    \midrule
    \midrule
    5  & 2.0 & 32 & 22.34 & 0.6411 \\
    15 & 2.0 & 32 & 23.19 & 0.6665 \\
    35 & 2.0 & 32 & 22.68 & 0.6527 \\
    15 & 1.0 & 32 & 22.97 & 0.6631 \\
    15 & 4.0 & 32 & 23.33 & 0.6624 \\
    15 & 2.0 & 48 & 22.53 & 0.6482 \\
    15 & 2.0 & 64 & 21.13 & 0.6351 \\
    \bottomrule
  \end{tabular}
  \label{tab:3}
\end{table}

As shown in Table~\ref{table:2}, our RBDM outperforms state-of-the-art methods on the NTIRE2020, NTIRE2021, and NTIRE2023 datasets, achieving notable improvements in PSNR (e.g., +3.67 dB) and SSIM (e.g., +0.0177). Visual comparisons in Fig.~\ref{fig:4} highlight the superiority of our method over existing approaches. Our RBDM excels in restoring fine details and producing more natural images in light haze conditions, leading to higher PSNR and SSIM scores. However, its performance diminishes in heavy haze, where fine details are often obscured. 
In such cases, methods with more aggressive haze removal may still produce acceptable results despite lower PSNR and SSIM. 

\subsection{Ablation Study}

% The whole formulation of our proposed model is methodically crafted based on the strategy of residual shifting. It is thus impossible to make an ablation study. However, we have provided sufficient ablation analysis on the other configurations an shown in Table~\ref{tab:3}.
\begin{figure}[t]
\centering
\includegraphics[width=0.9\columnwidth]{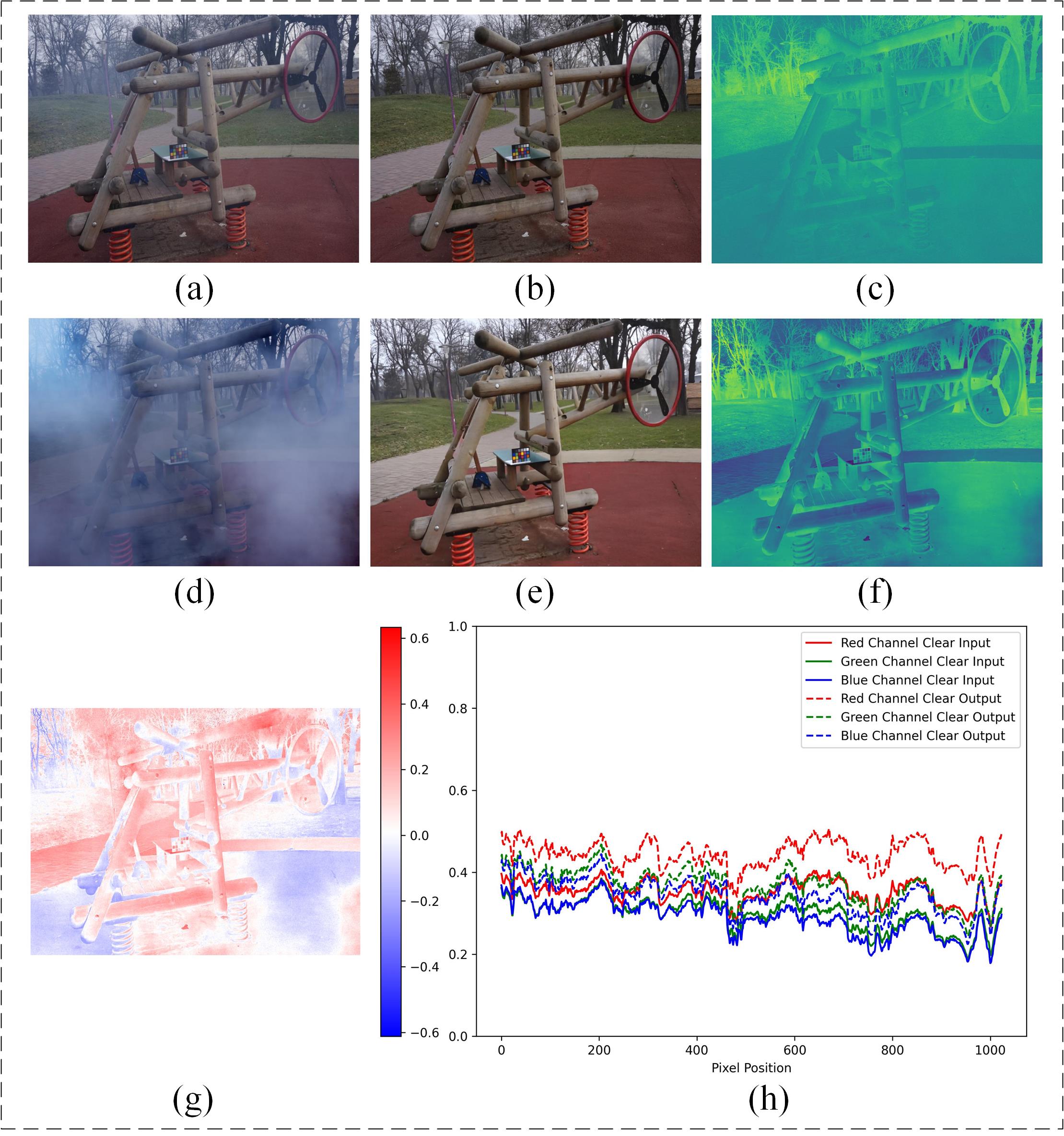} 
\caption{Evaluation of RBDM Effectiveness. (a) Hazy input. (b) Clear input. (c) Haze map for hazy input. (d) Hazed output of the clear input. (e) Dehazed output of the hazy input. (f) Haze map for hazy output. (g) Difference in haze distribution. (h) Average RGB channel intensity curves.}
\label{fig:5}
\end{figure}

The effectiveness of the proposed method is demonstrated through the visual analysis in Fig.\ref{fig:5}. Subfigures (c) and (f) show haze maps for the hazy input (a) and the hazed output (d) generated from the clear reference (b), with brightness indicating haze density. Subfigure (g) highlights the difference in haze distribution between the initial hazy input and the artificially hazed output, illustrating the model's ability to generate realistic, non-uniform haze patterns. This suggests that our method is effective for dataset augmentation of both hazy and haze-free images. Subfigure (h) compares the RGB channel intensity curves of the clear reference (b) and dehazed output (e), where their close alignment confirms the method's ability to preserve color fidelity, ensuring that dehazed images retain their natural appearance.

\section{CONCLUSION}

In this paper, we present RBDM, a residual-based efficient bidirectional diffusion model that successfully realizes both image dehazing and haze generation with individual timesteps. Unlike traditional models that reconstruct toward Gaussian noise, RBDM reconstructs hazy-to-haze-free and haze-free-to-hazy images, shortening the diffusion length and leveraging prior information more effectively. By employing a unified score function on image patches, RBDM enables size-agnostic bidirectional transitions between haze-free and hazy images. Extensive experiments on synthetic and real-world datasets validate the method’s effectiveness. Future work will focus on recovering details in heavily hazed regions and improving robustness across varying haze conditions.

\bibliographystyle{IEEEbib}
\bibliography{icme2025references}

\end{document}